\documentclass[lettersize,journal]{IEEEtran}
\usepackage{amsmath,amsfonts}
\usepackage{algorithmic}
\usepackage{algorithm}
\usepackage{array}
\usepackage[caption=false,font=normalsize,labelfont=sf,textfont=sf]{subfig}
\usepackage{textcomp}
\usepackage{stfloats}
\usepackage{url}
\usepackage{verbatim}
\usepackage{graphicx}
\usepackage{cite}
\usepackage{hyperref}
\usepackage{multirow}
\usepackage{pifont}
\usepackage{caption}
\usepackage{booktabs}
\usepackage{todonotes}

\begin{document}

\title{CaBuAr: California Burned Areas dataset for delineation}

\author{Daniele Rege Cambrin, Luca Colomba, Paolo Garza \thanks{Daniele Rege Cambrin, Luca Colomba and Paolo Garza are with Politecnico di Torino, Turin, Italy}}



\maketitle

\begin{abstract}
Forest wildfires represent one of the catastrophic events that, over the last decades, caused huge environmental and humanitarian damages. In addition to a significant amount of carbon dioxide emission, they are a source of risk to society in both short-term (e.g., temporary city evacuation due to fire) and long-term (e.g., higher risks of landslides) cases. Consequently, the availability of tools to support local authorities in automatically identifying burned areas plays an important role in the continuous monitoring requirement to alleviate the aftereffects of such catastrophic events.
The great availability of satellite acquisitions coupled with computer vision techniques represents an important step in developing such tools. This paper introduces a novel open dataset that tackles the burned area delineation problem, a binary segmentation problem applied to satellite imagery. The presented resource consists of pre- and post-fire Sentinel-2 L2A acquisitions of California forest fires that took place starting in 2015. Raster annotations were generated from the data released by California's Department of Forestry and Fire Protection. Moreover, in conjunction with the dataset, we release three different baselines based on spectral indexes analyses, SegFormer, and U-Net models.
\end{abstract}


\section{Introduction}
The Earth Observation (EO) field has greatly increased the number of applications in the last decades thanks to the greater data availability, storage capacity, and computational power of modern systems. In fact, leveraging data acquired by Sentinel, Landsat, and Modis missions as an example, it is possible to retrieve information at a continental scale in a short amount of time. This, in conjunction with the development of modern methodologies in the field of machine learning and deep learning, represents an extremely interesting area of research for scientists and authorities from different fields, such as governments and first responders involved in disaster response and disaster recovery missions. Phenomena such as climate change and extreme climate events have a tremendous societal, economic, and environmental impact, also leading to humanitarian and environmental losses (e.g., higher risk of landslide due to a forest fire). Indeed, leveraging EO and modern deep learning methodologies can provide useful tools in the area of disaster management and disaster recovery. Within the research community, numerous previous works proved the effectiveness of computer vision architectures in the field of disaster response, such as flood delineation~\cite{dong2021monitoring}, %flood_segmentation
change detection~\cite{asokan2019change, sublime2019automatic} and burned area delineation~\cite{lasaponara2019identification, cambrin2022vision, tanase2020burned}.
This paper fits in the last mentioned context. Specifically, we release a dataset to tackle the burned area delineation problem, i.e., a binary image segmentation problem that aims to identify areas damaged by a forest wildfire. Tackling such a problem with modern methodologies requires great data availability. However, the time and cost needed to produce high-quality annotations severely limit the ability to investigate ad-hoc solutions in the EO field.
For these reasons, we propose a new dataset related to forest fires in California, collecting data from the Sentinel-2 mission, which has been specifically designed to monitor land surface changes~\cite{esa_sentinel2}. % that took place after the year 2015. 
The dataset is publicly available to the research community at \url{https://huggingface.co/datasets/DarthReca/california_burned_areas}.

Ground truth masks for the task of binary image segmentation were generated starting from the public vector data provided by California's Department of Forestry and Fire Protection~\cite{calfire} and rasterized. Satellite acquisitions, i.e., the raw input data, were instead collected from the Sentinel-2 L2A mission through Copernicus Open Access Hub. More precisely, we collected and released both pre-fire and post-fire information associated with the same area of interest.

The contributions of this paper can be summarized as follows:
\begin{itemize}
    \item A novel image segmentation dataset consisting of Sentinel-2 pre- and post-fire acquisitions. Thus, the task can be addressed with a twofold approach: image segmentation task and change detection task;
    \item Three different baselines were evaluated on the proposed dataset: one consisting of the evaluation of several burned area indexes and the Otsu's automatic thresholding method~\cite{otsu}, one based on the SegFormer model~\cite{segformer}, and one based on the U-Net model~\cite{unet}.
\end{itemize}
The paper is structured as follows. Section~\ref{sec:related_works} introduces the related work; Section~\ref{sec:dataset} introduces the collected dataset and the preprocessing steps performed, whereas Section~\ref{sec:task} and Section~\ref{sec:experiments} formally introduce the task and the experimental settings and results. Finally, Section~\ref{sec:conclusion} concludes the paper.

\section{Related Works}
\label{sec:related_works}
Before the development of deep learning-based methodologies, domain experts based their analyses on satellite imagery leveraging spectral index computation and evaluation. Considering the SAR context, thresholding-based techniques have been adopted to distinguish between flooded and unflooded areas~\cite{sen1floods11}. Different analyses have been performed on various tasks concerning several spectral indexes such as in cloud detection (cloud mask)~\cite{cloud_mask}, water presence (WP, NDWI)~\cite{bais2, water_indexes} and vegetation analysis (NDVI)~\cite{ndvi}. 

Considering the burned area delineation problem, domain experts have developed several indexes: NBR, NBR2, BAI, and BAIS2~\cite{bais2}. They are computed using different spectral bands to generate an index highlighting the affected areas of interest. Such techniques are often coupled with thresholding methodologies: either fixed or manually calibrated threshold values are chosen~\cite{calibrated_dnbr}, or automatic thresholding algorithms are used~\cite{otsu_burned_area}.
Additional studies evaluate index-based techniques with additional in-situ information, namely the Composite Burned Area Index (CBI), which indeed provides insightful information but does not represent a scalable solution because in-situ data are incredibly costly to collect. Furthermore, studies confirmed that finding a unique threshold that is region- and vegetation-independent is difficult~\cite{cansler2012robust}.

More recently, researchers started adopting supervised learning techniques to solve several tasks in computer vision and EO. More precisely, CNN-based models proved their effectiveness in image classification and segmentation tasks, achieving state-of-the-art performances compared to index-based methodologies~\cite{knopp2020deep, farasin2020supervised}.
The main drawback is the need for a significant amount of labeled data, possibly covering heterogeneous regions with different morphological features to learn better representations.
Over the years, many of the proposed frameworks limit their analyses to a few samples collected from a limited number of countries or locations~\cite{safder2022burnt}. % hu2021unitemporal

In the EO domain, different public datasets are available to the research community tackling different problems, such as flood delineation~\cite{sen1floods11}, %mmflood
% deforestation~\cite{amazon_rainforest_dataset_paper} and 
crop classification and segmentation~\cite{crop_classif_dataset} but, to the best of our knowledge, only two public datasets are available for the burned area delineation problem covering some countries in Europe~\cite{colomba2022cikm} and Indonesia~\cite{indonesia_dataset}. The dataset proposed in this paper collects both pre- and post-fire Sentinel-2 L2A data from California forest fires, limiting seasonal and phenological differences between the two acquisitions as explained in the following paragraphs. Table~\ref{tab:dataset_comparison} shows a comparison between the three datasets.

\begin{figure}
    \centering
    \includegraphics[width=.45\linewidth]{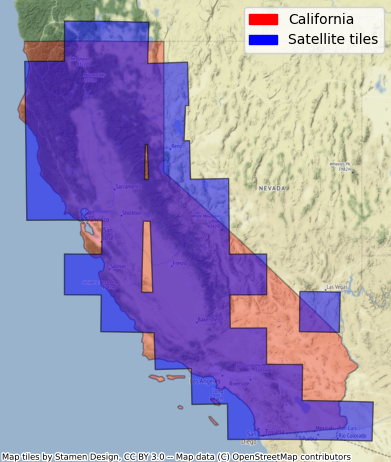}
    \caption{California administrative boundaries (red) vs satellite tiles of the proposed dataset (blue).}
    \label{fig:california_coverage}
    \vspace{-5pt}
\end{figure}

The proposed dataset consists of a higher number of satellite acquisitions (340 vs 73 vs 227) in California and a higher covered surface (Figure~\ref{fig:california_coverage}). Images are larger in terms of pixels (5490 vs 5000 (max) vs 512) and disclosed as raw data, as directly provided by Copernicus service. On the other hand, the European dataset provides data collected from a third-party service, for which the preprocessing operations are unknown. Furthermore, the monitored range of dates of the new dataset spans from 2015 to 2022, whereas the other two datasets span a smaller time period. 

\begin{table}[ht]
    \caption{Comparison between datasets. TD is the time difference between pre-fire and post-fire acquisitions.}
    \label{tab:dataset_comparison}
    \resizebox{\linewidth}{!}
    {
        \begin{tabular}{l|lll}
        \toprule
         & CaBuAr (ours) & \cite{colomba2022cikm} & \cite{indonesia_dataset} \\
        \midrule
        Region & California & Europe & Indonesia \\
        Mission & Sentinel-2 & Sentinel-1/2 & Landsat-8 \\
        Resolution & 20m & 10m (S2) & 30m \\
        Image size & 5490 $\times$ 5490 & up to 5000 $\times$ 5000 & 512 $\times$ 512 \\
        Raw data & \ding{51} & \ding{55} & \ding{51} \\
        Channels & 12 & 12 & 8 \\
        Forest Fires & 340 & 73 & 227 \\
        Start date & Jan, 2015 & July, 2017 & Jan, 2019 \\
        End date & Dec, 2022 & July, 2019 & Dec, 2021 \\
        Total surface & $\sim$450 000 km$^2$ & $\sim$19 000 km$^2$ & $\sim$46 000 km$^2$ \\
        Post-fire & \ding{51} & \ding{51} & \ding{51} \\
        Pre-fire & \ding{51} & \ding{51} & \ding{55} \\
        TD & $\sim$1 year & $\leq$ 2 months & / \\
        \bottomrule
        \end{tabular}
    }
    \vspace{-5pt}
\end{table}

\section{Dataset}
\label{sec:dataset}
\begin{figure}
    \centering
    \includegraphics[width=0.65\linewidth]{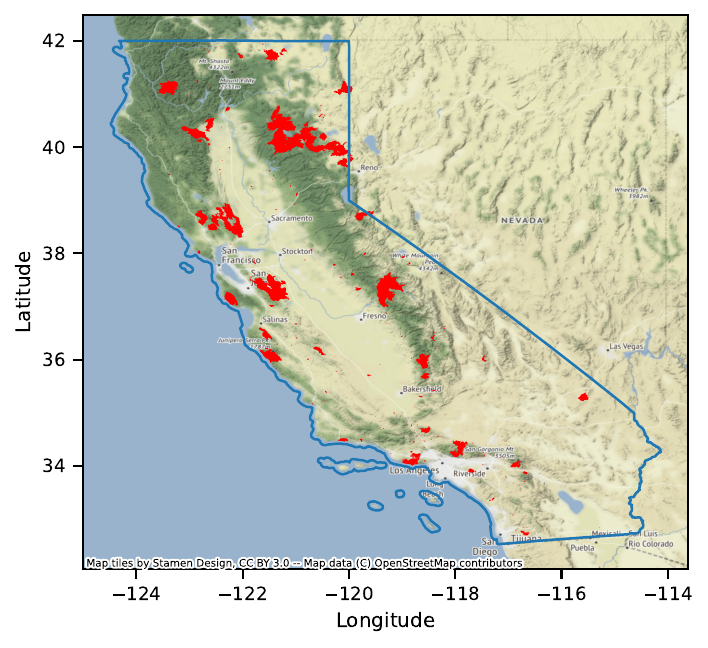}
    \caption{Geographical distribution of wildfires (red) within the California boundaries (blue).}
    \label{fig:fires_map}
    %\vspace{-15pt}
\end{figure}

The newly created dataset comprises L2A products of Sentinel-2, a European Space Agency (ESA) mission. The area of interest is California, with the geographical distributions of the events shown in Figure~\ref{fig:fires_map}. We collected images of the same area before and after the wildfire. It is essential to note that the L2A product contains RGB channels and other spectral bands in the infrared region and ultra blue for a total of 12 channels. Depending on the band, they have a resolution of $10m$, $20m$, or $60m$ per pixel.

\subsection{Preprocessing}

The California Department of Forestry and Fire Protection publicly provides the ground truth vector information, which we converted into raster images. Each pixel contains a binary value: 1 in the burned area and 0 in the case of undamaged areas. Although the registered wildfires span from 1898 to 2022, we collected data only for wildfires from 2015 onwards because there were no Sentinel-2 images before 2015. We gathered the Sentinel-2 images directly from Copernicus Open Access Hub.
%Although the registered wildfires span from 1950 to 2022, due to image availability of Sentinel-2 images, we collected data from wildfires that took place from 2015 onwards directly from Copernicus Open Access Hub.

% All L2A products were collected directly from Copernicus Open Access Hub and some preprocessing steps are necessary before making each product usable more conveniently.

Post-wildfire images are collected within one month after the containment date. A total number of 340 acquisitions were downloaded, each being of size $5490 \times 5490$ pixels with a resolution of $20m$ per pixel. Sentinel-2 bands with different resolutions were either upsampled or downsampled with bicubic interpolation to reach the target resolution.

Pre-fire images have the same size and resolution as the post-fire acquisitions. To enforce coherence and similar seasonal and phenological conditions, we downloaded pre-fire data considering a temporal window of 4 weeks, centered one year before the date post-fire data were collected. For example, given a post-fire acquisition collected in 2018/04/01, we downloaded the products available between 2017/03/18 and 2017/04/15, with center 2017/04/01. This ensures similar climatic and seasonal conditions, thus limiting environmental changes as much as possible. 
% {\bf Paolo. Toglierei la frase seguente}This was done because vast wildfire phenomena may last several months.{\bf toglierei fino a qui.}
%The date of these images is one year before the post-fire images in a window of two weeks before and after.
In some cases, retrieving these products was impossible due to data unavailability, i.e., not all wildfires have a pre-fire acquisition satisfying such a constraint. Given the 340 wildfires considered in this study, 208 have pre-fire availability satisfying the abovementioned constraint.

The dataset was randomly split into five non-overlapped folds with the positive class (burned area) equally distributed to perform cross-validation.
% The dataset was split randomly into 5 folds to balance the possibilities and make it usable for cross-validation. 

\begin{figure}
    \centering
    \includegraphics[width=.85\linewidth]{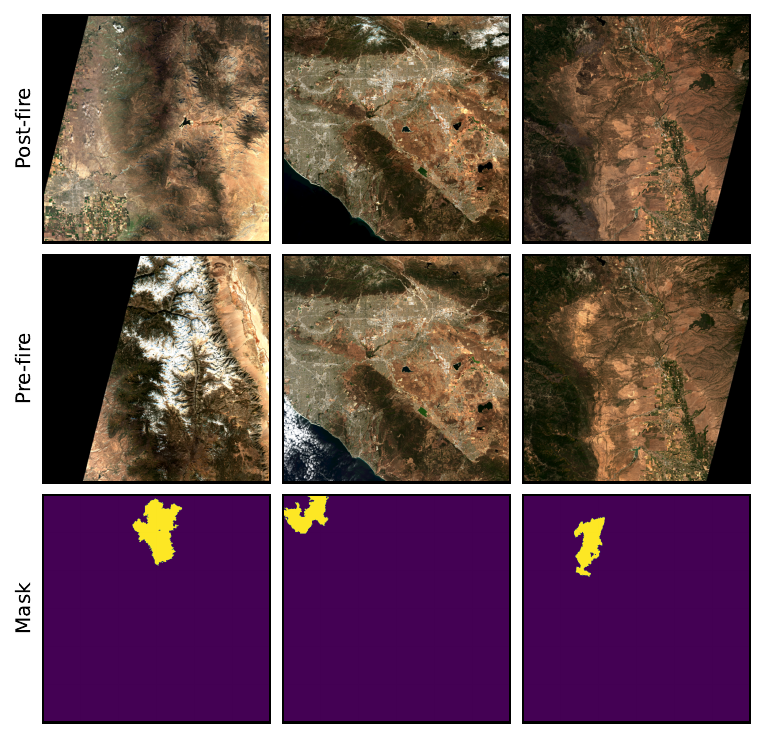}
    \caption{Example of pre-fire and post-fire RGBs and relative masks.}
    \label{fig:pre_post_examples}
    \vspace{-17pt}
\end{figure}

% Due to the size of the original input images (variable, up to $5490 \times 5490$), we split them into patches of size $512 \times 512$. Furthermore, due to class imbalance, we decided to keep only those patches containing at least one pixel associated with the positive class and no clouds over the area of interest. %(comment 6).

% {\bf Paolo. Forse la descrizione della divisione in patch 512x512 va riportata qui con anche il numero di quante sono queste patch.}

\subsection{Manual inspection}

After collecting data, we manually evaluated each post-fire image using RGB channels. This was done to (i)~discard invalid samples and (ii)~ enrich the dataset with metadata and comments based on our subjective evaluation. We remark that such comments are not helpful for the final prediction task but can be used to better characterize the data. 
Our evaluation is associated with each satellite acquisition. % and was obtained by inspecting smaller patches of size 512 $\times$ 512 which involved the affected areas.
% Our evaluation is unique for each sample and refers to all patches of size $512 \times 512$ extracted from each image as described in \ref{sec:experimental_settings}.

\begin{figure}
    \centering
    \includegraphics[width=\linewidth]{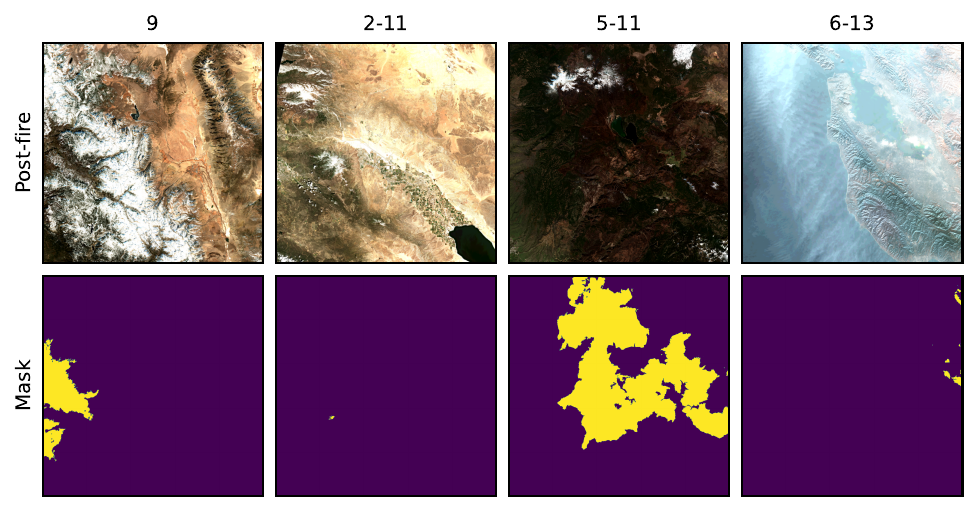}
    \caption{Sample of post-fire RGBs and masks with the associated comments.}
    \label{fig:example_cases}
    \vspace{-15pt}
\end{figure}

Each image has a metadata field with a list of numeric codes generated from the manual inspection. Figure~\ref{tab:comments_meaning} reports the code-to-comment association. As can be seen, different climatic conditions can be found in the dataset. 
%In \ref{fig:example_cases}, some examples of post-fire images composed by RGB channels, relative masks and one of the comments assigned. 
Figure~\ref{fig:example_cases} reports some examples of post-fire images. For each post-fire acquisition,  Figure~\ref{fig:example_cases} reports its RGB version (first line), its binary mask (second line), and the comment(s) assigned to it (on top of the RGB image). For instance, the second acquisition has two comments: 2 and 11.
We noted that some masks seem to overestimate the burned area. However, our perception refers to the RGB version of the images, i.e., to a subset of the available information. Moreover, our subjective perception can be biased also because the regions at the borders of burned areas are usually less damaged than the central ones.
%We noted that some masks seem to overestimate the burned area, but our perception refers to the RGB only and probably areas at the borders demonstrated negligible to slight damage compared to central areas. 
These notes must be extended to other mask-related comments, but they are rarer. Comments are equally distributed among the folds. All the images marked with comments that can negatively affect results and the dataset's quality (i.e., 4, 8, and 12) were discarded.

Finally, each pre-fire image was manually inspected to verify its validity, but no new comment types were added. All invalid pre-fire acquisitions were discarded.

\begin{table}[]
    \caption{Association between code and comment.}
    \label{tab:comments_meaning}
    \centering
    \begin{tabular}{c|c}
        \textbf{comment} & \textbf{meaning} \\
        \cline{1-2}
        0 & Affected area is in the incomplete region\\
        \cline{1-2}
        1 & Image is incomplete \\
        \cline{1-2}
        2 & Small burned area \\
        \cline{1-2}
        3 & Mask has a small offset \\
        \cline{1-2}
        4 & Mask is totally wrong \\
        \cline{1-2}
        5 & Extensive burned area \\
        \cline{1-2}
        6 & Clouds over the burned area \\
        \cline{1-2}
        7 & Too many clouds over the image \\
        \cline{1-2}
        8 & Wildfire ongoing \\
        \cline{1-2}
        9 & Snow on the burned area \\
        \cline{1-2}
        10 & Mask seems smaller than the burned area \\
        \cline{1-2}
        11 & Mask seems bigger than the burned area \\
        \cline{1-2}
        12 & Mask is in the missing data area \\
        \cline{1-2}
        13 & Part of the mask is outside the area \\
        \cline{1-2}
    \end{tabular}
    \vspace{-5pt}
\end{table}

% COMMENT DISTRIBUTION
% \begin{figure}
%     \centering
%     \includegraphics[width=\linewidth]{images/graphs/comments.pdf}
%     \caption{Comments distribution}
%     \label{fig:comments_distribution}
% \end{figure}

\section{Task}
\label{sec:task}
The proposed dataset can be used as a benchmark for different tasks in supervised and unsupervised scenarios: (i)~binary segmentation based on post-fire acquisitions only, (ii)~binary segmentation based on pre-fire and post-fire acquisitions, (iii)~binary change detection on pre-fire and post-fire images, and (iv)~performance assessment of spectral indexes.

Tasks~(i) and (ii) involve developing and training supervised learning algorithms to perform pixel-level prediction based on one or two acquisitions, respectively. Pixels are labeled either as burned or undamaged.

Task~(iii) can leverage unsupervised learning to extract differences between the two products and eventually exploit the ground truth to focus the task on detecting changes due to a wildfire. 

Instead, Task~(iv) may be of greater interest to domain experts in developing new spectral indexes designed for burned area identification. As such, researchers may use the dataset to evaluate the quality of the index by computing the separability index (SI)~\cite{bais2}.
% , defined as:

% \begin{equation}
%     SI = \frac{|\mu_b - \mu_u|}{\sigma_b + \sigma_u}
% \end{equation}

% \noindent being $\mu_b, \mu_u, \sigma_b, \sigma_u$ the average values and standard deviations of burned and undamaged pixels.

\section{Experiments}
\label{sec:experiments}
Our experiments test various classical and deep learning methods for three different settings:
\begin{enumerate}
    \item Usage of all the available post-fire images (Setting 1); \label{set:1}
    \item Usage of the subset of post-fire images for which the pre-fire data is available (Setting 2); \label{set:2}
    \item Usage of post-fire and pre-fire images. Thus, two input images are provided. Spectral indexes in this setting were evaluated by computing the difference between pre-fire and post-fire indexes (Setting 3). \label{set:3}
\end{enumerate}
The code for the experiments can be found at \url{https://github.com/DarthReca/CaBuAr}.

\subsection{Experimental settings}
\label{sec:experimental_settings}
The encoder of SegFormer is initialized with the original weights for Image-Net duplicated four times to handle the 12 available channels for Sentinel-2 L2A acquisitions. U-Net is instead randomly initialized. The batch size was set to 8. We used the AdamW optimizer with an initial learning rate of 0.001, decreased by a factor of 10 every 15 epochs, and a weight decay of 0.01 for every considered model. We used the well-known Dice loss~\cite{diceloss} as the loss function. All models were trained on a single Tesla V100 32GB GPU. The testing was made using the weights associated with the best validation loss.

Due to the size of the original input images ($5490 \times 5490$), we split them into patches of size $512 \times 512$. Furthermore, due to class imbalance, we kept only those patches containing at least one pixel associated with the positive class and no clouds over the area of interest (comment 6).
% We kept all the patches containing at least one burned pixel because we know in which area a wildfire happened before making its delineation.
% We excluded all images marked with comment 6 (this means some clouds are over the burned area). 
A total of 534 patches for setting~\eqref{set:1} and 356 for settings \eqref{set:2} and \eqref{set:3} were obtained.

The statistics and performances reported in the remainder of the paper refer to the data obtained after the split-and-filter process mentioned earlier. All training and evaluation procedures were performed with a cross-validation approach. The same criterion was also applied for spectral indexes methodologies to obtain comparable results, despite the absence of a trainable model. The reported values are expressed as mean and standard deviation computed over the five folds.

\begin{figure}[htb]
    \centering
    \subfloat[Setting \eqref{set:1}.]{\includegraphics[width=0.4\linewidth]{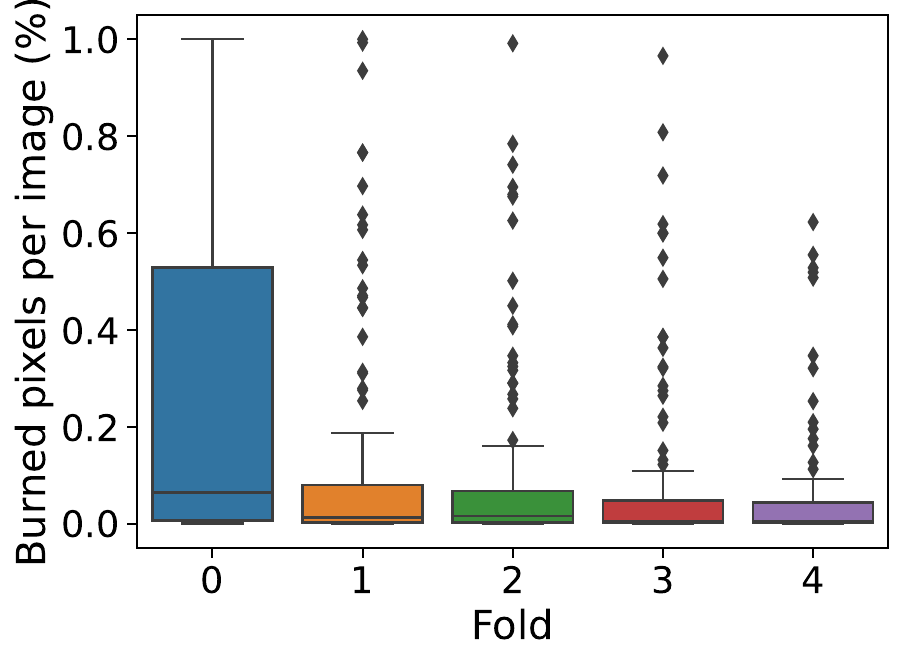}}
    \subfloat[Settings \eqref{set:2} and \eqref{set:3}.]{\includegraphics[width=0.4\linewidth]{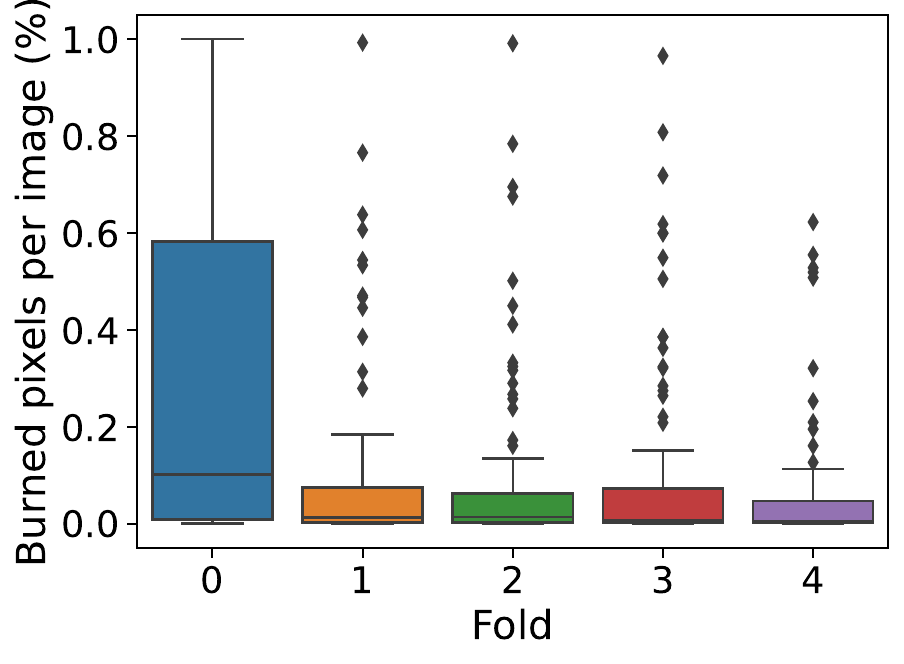}}
    \caption{Burned pixels percentage per image per fold.}
    \label{fig:burned_distribution}
    \vspace{-3pt}
\end{figure}

In Figure~\ref{fig:burned_distribution}, we highlight the distribution of burned pixels percentage per image in each fold. The problem is unbalanced, but the different folds share a similar distribution.

\subsection{Spectral Indexes}
We evaluated several spectral indexes (NBR, NBR2, BAI, BAIS2) for the burned area delineation task. 
%We report their definitions in terms of Sentinel-2 bands for clarity in \eqref{eq:nbr,eq:nbr2,eq:bai,eq:bais2}.

% \begin{equation}
%     \label{eq:nbr}
%     NBR = \frac{B8 - B12}{B8 + B12}
% \end{equation}
% \begin{equation}
%     \label{eq:nbr2}
%     NBR2 = \frac{B11 - B12}{B11 + B12}
% \end{equation}
% \begin{equation}
%     \label{eq:bai}
%     BAI = \frac{1}{(0.1 - B4)^2 + (0.06 - B8)^2}
% \end{equation}
% \begin{equation}
%     BAIS2 = \bigg(1 - \sqrt{\frac{B6 \cdot B7 \cdot B8A}{B4}}\bigg) \bigg(\frac{B12 - B8A}{\sqrt{B12 + B8A}} + 1\bigg)
%     \label{eq:bais2}
% \end{equation}

\begin{figure}
    \centering
    \includegraphics[width=0.8\linewidth]{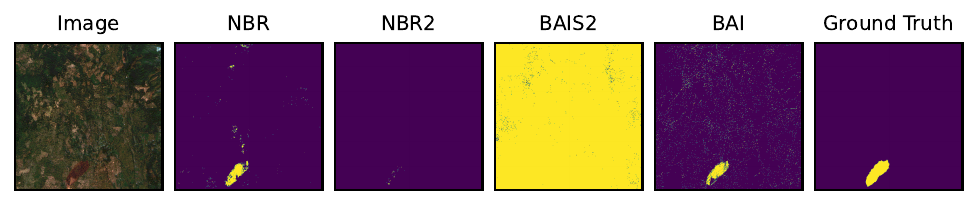}
    \caption{Example of segmentation using Otsu's method for different indexes.}
    \label{fig:example_otsu}
\end{figure}

% that are expressed by following formulas in terms of Sentinel-2 bands:
% \begin{equation}
% \end{equation}
In particular, to assess the performances on the dataset, we computed the Separability Index (visible in Table~\ref{tab:indexes_result}), which quantifies how well the index under analysis discerns between burned and unburned regions, i.e., a higher value of the SI implies that, in general, damaged areas have a different distribution of the spectral index. 
% We tested different spectral indexes (NBR, NBR2, BAI, BAIS2) for the task and reported the Separability Index in \ref{tab:sep_indexes}.
% They will probably perform poorly because their values are close to 0.
We apply Otsu's method to quantify the indexes' segmentation performances. Results are shown in Table~\ref{tab:indexes_result}, which confirm the poor performances in terms of F1-Score and IoU. Additional pre-fire information (Setting \eqref{set:3}) does not significantly improve the evaluation metrics. 
Figure~\ref{fig:example_otsu} shows an example of predictions for the cited indexes and Otsu. In this case, BAI and NBR achieve better scores, but many disturbances affect the final result in the unburned regions.

% \begin{table}
%     \centering
%     \begin{tabular}{c|ll}
%         \toprule
%         Setting & Index & SI \\
%         \midrule 
%         \multirow{4}*{\eqref{set:1}} & NBR &  \textbf{0.294} \\
%         & NBR2 & 0.224 \\
%         & BAI & 0.044 \\
%         & BAIS2 & 0.027 \\
%         \hline
%         \multirow{4}*{\eqref{set:2}} & NBR &  0.320 \\
%         & NBR2 & \textbf{0.349} \\
%         & BAI & 0.052 \\
%         & BAIS2 & 0.002 \\
%         \hline
%         \multirow{4}*{\eqref{set:3}} & dNBR &  \textbf{0.247}\\
%         & dNBR2 & 0.189\\
%         & dBAI & 0.040\\
%         & dBAIS2 & 0.027\\
%     \end{tabular}
%     \caption{Separability Indexes (SI) computed for each setting and each evaluated index.}
%     \label{tab:sep_indexes}
% \end{table}

\begin{table}
    \caption{Separability Indexes (SI) and metrics computed for each setting and each evaluated index.}
    \label{tab:indexes_result}
    \centering
    \begin{tabular}{c|llll}
        \toprule
        Setting & Index & SI & F1 Score & IoU \\
        \midrule 
        \multirow{4}*{\eqref{set:1}} 
        & NBR &  \textbf{0.294} & 0.150±0.231 & 0.103±0.180 \\
        & NBR2 & 0.224 & \textbf{0.226±0.269} & \textbf{0.159±0.209} \\
        & BAI & 0.044 & 0.040±0.121 & 0.026±0.086 \\
        & BAIS2 & 0.027 & 0.194±0.292 & 0.148±0.252\\
        \hline
        \multirow{4}*{\eqref{set:2}} 
        & NBR &  0.320 & 0.106±0.196 & 0.071±0.150 \\
        & NBR2 & \textbf{0.349} & \textbf{0.243±0.278} & \textbf{0.172±0.218} \\
        & BAI & 0.052 & 0.037±0.115 & 0.024±0.079 \\
        & BAIS2 & 0.002 & 0.086±0.174 & 0.057±0.138 \\
        \hline
        \multirow{4}*{\eqref{set:3}} 
        & dNBR &  \textbf{0.247} & 0.114±0.212 & 0.079±0.168 \\
        & dNBR2 & 0.189 & \textbf{0.218±0.281} & \textbf{0.157±0.225} \\
        & dBAI & 0.040 & 0.066±0.161 & 0.045±0.127 \\
        & dBAIS2 & 0.027 & 0.047±0.126 & 0.030±0.099 \\
    \end{tabular}
    \vspace{-10pt}
\end{table}

\subsection{Deep learning models}

We tested two deep learning architectures for semantic segmentation: a CNN (U-Net~\cite{unet}) and a Vision Transformer (SegFormer~\cite{segformer}). We decided to take into account two different versions of SegFormer (\textit{B0}, the smallest version, and \textit{B3}, a mid-range version) that differ only in size and so in the number of parameters. U-Net, SegFormer-B0, and SegFormer-B3 consist of 31M, 3.8M, and 47M parameters, respectively. 
To deal with Setting~\eqref{set:3}, the two input images (pre- and post-fire) are concatenated along the channel axis, creating patches of size $24 \times 512 \times 512$ ($C \times H \times W$).
We reported the results for the different settings and models in Table~\ref{tab:deep_learning_results}.

Without any specific pre-training, U-Net provides in every setting the best performance. SegFormer-B0, which is also lighter than U-Net, provides comparable performance, having some difficulties only with Setting~\eqref{set:3}. SegFormer-B3 does not justify the greater complexity considering its results.

\begin{figure}[htb]
    \centering
    \includegraphics[width=0.8\linewidth]{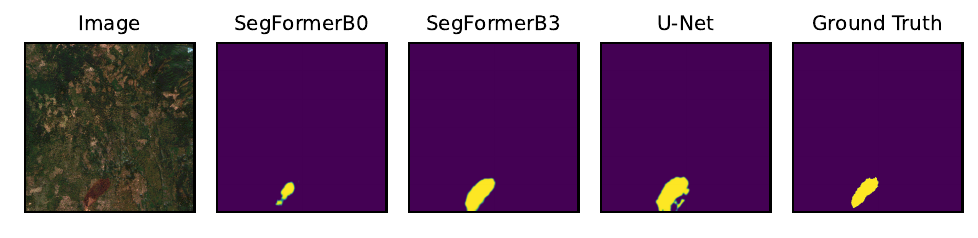}
    \caption{Examples of prediction with deep learning models.}
    \label{fig:examples_deep}
    \vspace{-13pt}
\end{figure}

Figure~\ref{fig:examples_deep} reports the predictions of these models on the same input sample shown in Figure~\ref{fig:example_otsu}. The most evident difference is deep models tend to be more precise and less affected by disturbances.

\begin{table}[]
    \caption{Metrics calculated for each deep learning model evaluated.}
    \label{tab:deep_learning_results}
    \centering
        \begin{tabular}{c|l|ccc}
            \toprule
            Setting & Metric & SegFormerB3 & SegFormerB0 & U-Net \\
            \midrule 
            \multirow{2}*{\eqref{set:1}} 
            & F1 Score & 0.620±0.009 & 0.686±0.004 & \textbf{0.707±0.004}\\
            & IoU & 0.497±0.008 & 0.563±0.004 & \textbf{0.583±0.004} \\
            \hline
            \multirow{2}*{\eqref{set:2}} 
            & F1 Score & 0.583±0.014 & 0.654±0.003 & \textbf{0.705±0.002} \\
            & IoU & 0.447±0.012 & 0.535±0.003 & \textbf{0.577±0.002} \\
            \hline
            \multirow{2}*{\eqref{set:3}} 
            & F1 Score & 0.533±0.003 & 0.499±0.009 & \textbf{0.625±0.002} \\
            & IoU & 0.401±0.003 & 0.370±0.007 & \textbf{0.494±0.002} \\
        \end{tabular}
    \vspace{-15pt}
\end{table}

\section{Conclusion}
\label{sec:conclusion}
This paper introduced a new dataset for burned area delineation containing samples with different morphological features and some of California's largest wildfires. The dataset includes both pre-fire and post-fire data. 

We provided baselines based on different approaches to encourage the investigation in various research areas, such as semantic segmentation and change detection. This publicly available dataset can benefit researchers and public authorities for many tasks, such as recovery planning, constant monitoring of affected areas, and developing deep learning models for burned area delineation. We plan to extend the dataset to new regions and different satellite acquisitions in future work, making the dataset multimodal. Additionally, we plan to release the dataset sampled to different resolutions. The collection of satellite acquisitions is made publicly available to encourage future use and research activities.

\bibliographystyle{IEEEtran}
\bibliography{IEEEabrv,references}

\vfill

\end{document}